\PassOptionsToPackage{numbers,compress}{natbib}
\documentclass{article}

\usepackage[dblblindworkshop, final]{neurips_2025}
\usepackage[utf8]{inputenc} 
\usepackage[T1]{fontenc}    
\usepackage{hyperref}       
\usepackage{url}            
\usepackage{booktabs}       
\usepackage{amsfonts}       
\usepackage{nicefrac}       
\usepackage{microtype}      
\usepackage{xcolor}         
\usepackage{graphicx}
\usepackage{amsmath}
\usepackage{algorithm}
\usepackage{algorithmic}
\usepackage{makecell}

\workshoptitle{Efficient Reasoning}

\title{SGD-KV: Summarization Guided KV Cache Compression}

\author{%
  Zeyu Liu\thanks{Work done during an internship at Amazon. Correspondence to: Srikanth Ronanki <ronanks@amazon.com>} \\
  USC, USA \\
  \And
  Woomin Song \\
  KAIST, Korea \\
  \And
  Xuandi Fu \\
  Amazon AGI, USA \\
  \And
   Sai Muralidhar Jayanthi \\
  Amazon AGI, USA \\
  \And
  Vivek Govindan \\
  Amazon AGI, USA \\
  \And
  Aram Galstyan \\
  Amazon AGI, USA \\
  \And
  Sravan Babu Bodapati \\
  Amazon AGI, USA \\
  \And
  Srikanth Ronanki \\
  Amazon AGI, USA \\
}

\begin{document}

\maketitle

\begin{abstract}

Large language models (LLMs) face severe memory bottlenecks in long-context inference due to the linearly growing size of key-value (KV) caches. Existing KV cache compression techniques typically rely on simple heuristics, overlooking the distinct functional roles of different attention heads. We present SGD-KV (Summarization-Guided KV Cache Compression), a head-aware framework that leverages a novel chunk-summarization diagnostic task to systematically identify and prioritize attention heads specialized in hierarchical information aggregation. Experiments on Qwen2.5-7B-1M and Qwen3-32B across diverse long-context benchmarks demonstrate that SGD-KV achieves state-of-the-art performance with contexts up to 1M tokens, while reducing KV cache memory usage by up to 75\%. Our findings show that strategically allocating the KV cache budget based on the summarization score distribution of attention heads yields a superior efficiency–accuracy trade-off for long-context inference.


\end{abstract}

\section{Introduction}
\vspace{-3mm}

The push towards million-token context windows in Large Language Models (LLMs) like Qwen2.5-1M~\citep{qwen2.5} and Gemini 2.5~\citep{gemini25} is severely hampered by a fundamental obstacle: the prohibitive memory cost of the Key-Value (KV) cache, which scales linearly with context length. 
While methods exist to compress the KV cache, from early token-level eviction to more recent head-aware approaches~\citep{fu2024not,feng2024ada}, they often rely on simple, retrieval-based importance metrics to allocate cache budget. 
These heuristics fall short in complex scenarios like multi-document analysis or long-form dialogue, which demand hierarchical information aggregation rather than simple pattern matching.
We argue that a specialized subset of attention heads, which we term “summarization heads,” are primarily responsible for this higher-order cognitive function. To leverage this insight, we introduce SGD-KV (Summarization-Guided KV Cache compression), a framework that systematically identifies and prioritizes these crucial heads. 
Using a novel chunk-summarization diagnostic task, SGD-KV scores each head’s summarization capability and applies a water-filling-inspired algorithm to intelligently allocate cache budget for each head. Validated on benchmarks including OpenAI Multi-Round Co-Reference Resolution (MRCR)~\citep{openai_mrcr} and ETHIC~\citep{ethic}, SGD-KV sets a new state-of-the-art (SOTA), reducing KV cache usage by up to 75\% on contexts up to 1 million tokens without compromising model accuracy.

\section{Related Work}
\textbf{KV Cache Compression}
Research on mitigating the memory burden of KV caches has progressed along two fronts: token-level eviction and head-level budget allocation. Early efforts like StreamingLLM~\citep{xiao2024sink} and H2O~\citep{zhang2023h2o} focused on identifying and discarding less important tokens based on heuristics like recency and cumulative attention scores. While reducing memory, these methods are functionally unaware. More sophisticated approaches like PyramidKV~\citep{cai2024pyramidkv} and AdaKV~\citep{feng2024ada} introduce dynamic cache allocation across different layers and heads, but their strategies are still guided by quantitative attention patterns rather than the semantic role of each head.

\textbf{Attention Head Specialization}
Concurrently, research into attention mechanism interpretability has demonstrated that 
heads often specialize. This was first shown for “retrieval heads” identified via needle-in-a-haystack tasks~\citep{wu2025retrieval}, and later for other pattern-matching roles like induction heads~\citep{tang2025razorattention} and retrieval-reasoning (R2) heads~\citep{fu2024not}. However, these functional discoveries have been driven by retrieval-centric diagnostics, overlooking heads specialized for higher-order cognitive tasks. Our work bridges this divide: we introduce a summarization-based task to identify heads that perform hierarchical information synthesis and, for the first time, leverage this functional insight to create a more intelligent and efficient KV cache compression strategy.
\begin{figure*}[t]
  \includegraphics[width=0.98\linewidth]{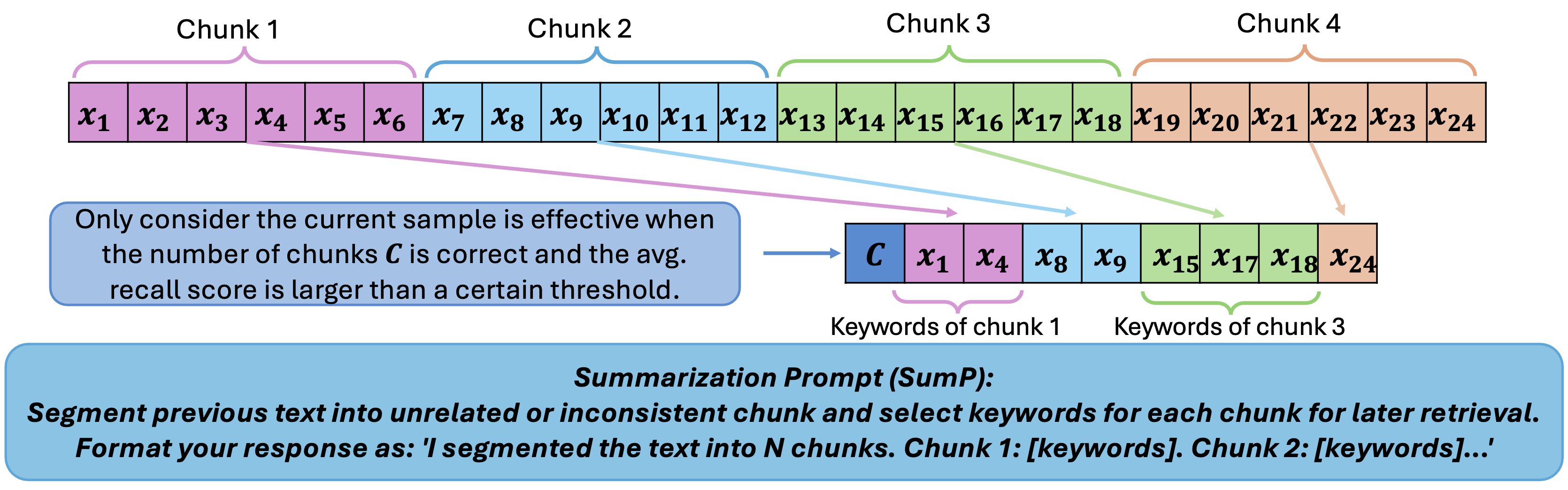}
  \caption{The illustration of chunk-summarization task.}
  \label{fig:overall}
\end{figure*}

\section{Method}

\subsection{Chunk-Summarization Task Design}
To effectively measure a head's ability to aggregate information, we design a diagnostic task that requires multi-level understanding, going beyond simple pattern matching. 
As illustrated in Figure~\ref{fig:overall}, we first construct long-context samples by concatenating multiple documents from various short-text summarization datasets (e.g., CNN/Dailymail dataset~\citep{cnndm}, DialogSum dataset~\citep{dialogsum}) and record the boundaries of each original document. 
Given a concatenated sample, we prompt the model to perform a two-part task:
(1) \textbf{Chunk Identification}: Identify the number of distinct semantic chunks within the concatenated document;
(2) \textbf{Keyword Extraction}: For each identified chunk, generate a concise list of keywords that capture its core meaning.
This design compels the model to first parse the document structure at a high level and then distill the semantic essence of each segment, providing a strong signal for identifying heads involved in hierarchical aggregation.

\subsection{Summarization Score Calculation}

To quantify each head's summarization capability, we first filter for valid model responses as shown in Fig.~\ref{fig:overall} and then compute an attention-based importance score.
For each valid sample, we compute an importance score $I_h$ for every attention head $h$. The score measures the strength of attention from the generated keywords back to their corresponding source text within the correct chunk.
\begin{equation}
I_h = \frac{1}{c} \sum_{m=1}^{c} \frac{1}{|\mathcal{K}_m|} \sum_{j \in \mathcal{K}_m} \max_{n} A_h(p_{j}^o, p_{j, n}^i)
\end{equation}
where $\mathcal{K}_m$ is the set of keywords for chunk $m$, $p_{j}^o$ is the position of an output keyword token $j$, and $p_{j, n}^i$  is the $n^{th}$ occurrence of that same keyword in the input text. The $\max$ operation identifies the strongest attention link from a generated keyword to its most salient source token, effectively capturing the head's ability to pinpoint and aggregate key information. The final summarization score for each head is its average importance score across all valid samples.


\subsection{KV Cache Budget Allocation}
The computed summarization scores directly guide our head-aware KV cache allocation strategy. Similar to HeadKV~\citep{fu2024not}, we first normalize the scores across all heads such that they sum to one, then we always preserve a fixed budget for initial sink tokens ($b_{\text{sink}}$) and a sliding window of recent tokens ($b_{\text{window}}$) as the instructions of users are usually in the beginning or end of the whole input in practice. 

The remaining cache budget for the middle, compressible portion of the sequence ($M = C - b_{\text{sink}} - b_{\text{window}}$, where $C$ is total sequence length) is then allocated to each head $h$ in proportion to its normalized summarization score $S_h$, at a given overall KV cache budget ratio $R$:
\begin{equation}
    b_h = (S_h\cdot L\cdot H\cdot R)\cdot M + b_{\text{sink}} + b_{\text{window}}
    \label{eq:allocation}
\end{equation}
where $L$ is the number of layers, and $H$ is the number of KV heads. 
This ensures that heads identified as crucial for summarization receive a larger cache budget, allowing them to retain more historical context. 
We further adapt the Water-filling algorithm to reallocate the excessive KV budget from heads with very high scores to other heads with comparatively high scores, as detailed in Appendix~\ref{sec:appendix:redist}.
Within each head's allocated budget $b_h$, we use a standard token selection mechanism based on cumulative attention scores, similar to SnapKV~\citep{li2024snapkv}, to select the most important KV  entries to keep. 



\section{Experiments}


\textbf{Experimental Setup} We evaluate our approach on Qwen2.5-7B-Instruct-1M ~\citep{qwen2.5} and Qwen3-32B ~\citep{qwen3technicalreport} as representative models for non-reasoning small LMs and reasoning-capable large LMs, respectively.
However, we observed that the Qwen2.5-7B-Instruct-1M checkpoint from HuggingFace exhibits poor performance on multi-turn conversation tasks, like the OpenAI Multi-Round Co-Reference Resolution (MRCR) benchmark~\citep{openai_mrcr}. 
To address this limitation, we perform additional fine-tuning using a diverse dataset. Detailed fine-tuning setup is provided in the Appendix~\ref{sec:appendix:sft}. We also demonstrate the generalization of summarization heads when using different summarization datasets and quantitatively compare the head score distribution with other types of heads in Appendix~\ref{sec:appendix:iou} and visualize the final KV cache allocation for different types of heads in Appendix~\ref{sec:appendix:viz}

\begin{table}[h]
  \centering
  \caption{Accuracy comparison on MRCR dataset (25\% KV cache budget for eviction methods).}
\begin{tabular}{lccccccccc}
\toprule
& \multicolumn{4}{c}{\textit{Qwen3-32B}} & \multicolumn{5}{c}{\textit{Qwen2.5-7B-Instruct-1M}} \\
\cmidrule(r){2-5}  \cmidrule(r){6-10} 
Method & 8k & 16k & 32k & 64k & 64k & 128k & 256k & 512k & 1M \\
\cmidrule(r){2-5}  \cmidrule(r){6-10} 
FullKV & 79.22 & 70.40 & 68.18 & 46.42 & 95.01 & 96.38 & 88.6 & 63.84 & 43.29 \\
Minference~\citep{jiang2024minference} & - & - & - & - & 97.11 & 95.43 & 82.21 & 57.39 & 43.87 \\
AdaKV~\citep{feng2024ada}       &  26.45 & 14.7 & 15.88 & 12.47 & 28.34 & 29.56 & 21.99 & 15.28 & 10.98 \\
DuoAttention~\citep{xiao2025duoattention} & 76.19 & 59.17 & 60.36 & 28.37 & \textbf{85.80} & \textbf{89.82} & 73.78 & 39.10 & 24.73 \\
HeadKV~\citep{fu2024not} & \textbf{78.16} & 65.90 & \textbf{65.72} & 34.90 & 68.52 & 74.81 & 66.98 & 44.36 & 28.90 \\
Ours & 77.05 & \textbf{68.22} & 62.02 & \textbf{40.13} & 85.09 & 87.19 & \textbf{83.29} & \textbf{48.86} & \textbf{34.16} \\
\bottomrule
  \end{tabular}
  \label{tab:mrcr:qwen25}
\end{table}

\noindent
\textbf{MRCR}
The MRCR benchmark~\citep{openai_mrcr} tests a model's ability to retrieve information from long, multi-turn dialogues. As shown in Table~\ref{tab:mrcr:qwen25}, On Qwen3-32B with Chain-of-Thought (CoT) reasoning, both SGD-KV and HeadKV significantly outperform DuoAttention, suggesting that fine-grained budget allocation is more effective than a binary classification of heads for complex reasoning tasks.
As for the fine-tuned Qwen2.5-7B-1M model, SGD-KV consistently outperforms other SOTA head-level allocation methods like HeadKV and AdaKV. While DuoAttention shows strong performance at shorter lengths, our fine-grained allocation strategy proves superior as the context grows, surpassing DuoAttention beyond 128K tokens and demonstrating the most robust performance at the 1M token scale. This highlights the increasing importance of nuanced head prioritization in ultra-long contexts.

\begin{table}[h]
\centering
\caption{Performance comparison on ETHIC benchmark. "AT", "OG", and "RC" represents the attributing, organizing  and recalling sub-tasks, and "Avg." represents the average performance.}
\begin{tabular}{lcccccccc}
\toprule             
& \multicolumn{4}{c}{\textit{Qwen2.5-7B-Instruct-1M}} & \multicolumn{4}{c}{\textit{Qwen3-32B}}  \\
\cmidrule(r){2-5}  \cmidrule(r){6-9} 
Method & AT  & OG & RC  & Avg. & AT  & OG & RC  & Avg. \\
\cmidrule(r){2-5}  \cmidrule(r){6-9} 
Full KV & 26.19 & 17.31 & 21.45 & 21.65 & 31.08 & 26.11 & 28.39 & 28.53 \\
DuoAttention~\citep{xiao2025duoattention} & 25.58 & \textbf{19.55} & 13.40 & 19.51 & 28.01 & 20.1 & 25.35 & 24.49 \\
HeadKV~\citep{fu2024not}       & 26.46 & 15.48 & 20.67 & 20.87 & 30.90 & \textbf{25.61} & 28.05 & 28.19 \\
SGD-KV        & \textbf{27.53} & 15.38 & \textbf{21.12} & \textbf{21.34} & \textbf{30.94} & 25.60 & \textbf{28.60} & \textbf{28.38}  \\
\bottomrule
\end{tabular}
\label{tab:ethic25}
\end{table}
\noindent
\textbf{ETHIC}
We also use the ETHIC benchmark~\citep{ethic}, which features tasks requiring high information coverage from the input context. Results in Table~\ref{tab:ethic25} show that SGD-KV achieves SOTA performance on both models. Notably, with only a 25\% KV cache, our method on Qwen3-32B (28.38 Avg.) performs comparably to the Full KV baseline (28.53 Avg.), closing the gap more effectively than any other method. The smaller performance margins between methods on ETHIC, compared to MRCR, are likely attributable to a performance ceiling imposed by the base models' capabilities on these highly complex tasks. Besides, we show the resutls on BABILong~\citep{babilong} benchmark in Appendix~\ref{sec:appendix:babi}.

\section{Ablation Study: Impact of Query on Token Selection}

\begin{table}[htbp]
  \centering
\caption{Accuracy comparison on MRCR dataset. Bold values indicate scores that surpass the corresponding Query-Aware baseline results from Table~\ref{tab:mrcr:qwen25}.}
\begin{tabular}{llcccccccc}
\toprule
Mode & Method & 8k & 16k & 32k & 64k & 128k & 256k & 512k & 1M \\
\midrule
\textbf{\textit{Query}} & HeadKV & \textbf{87.91}	& 64.4	& 54.08	& 48.73	& 52.62	& 41.13	& 26.21 & 20.38 \\
\textbf{\textit{Unaware}} & Ours   &  \textbf{93.5}	& \textbf{85.55}	& 73.02	& 70.94	& 77.56	& 71.06	& 44.38 & 33.20\\
\midrule
\textbf{\textit{Proxy}} & HeadKV & \textbf{90.48}	& 68.71	& 60.90	& 57.52	& 57.13	& 50.62	& 32.41 & 23.1\\
\textbf{\textit{Query}} & Ours   &  \textbf{97.19}	& \textbf{89.67}	& \textbf{82.90}	& 82.62	& 83.96	& 80.65	& \textbf{50.00} & 32.24\\
\bottomrule
\end{tabular}
\label{tab:query}
\end{table}

Our KV cache compression involves two stages: head-level budget allocation and token-level selection. The token selection step, similar to SnapKV~\citep{li2024snapkv}, uses the cumulative attention scores generated by a final observation window to identify important tokens. 
In this study, we ablate the choice of this observation window to understand its impact on performance. We define three conditions:
\textbf{Query-Aware} (Default): The standard approach used in our main results, where the last 128 tokens (inclduing the real query) guides token selection.
\textbf{Query-Unaware}: The final question is excluded, and the last 128 tokens of the preceding context are used to guide selection. This simulates a scenario where the specific query is unknown during compression.
\textbf{Proxy-Query}: A fixed, task-agnostic prompt—{\textit{"Segment previous text into unrelated or inconsistent chunk and select keywords for each chunk for later retrieval." }}—is used as a universal proxy for the final query.

The results presented in Table~\ref{tab:query} illustrates that removing the final question's guidance causes a significant performance degradation for both SGD-KV and HeadKV compared to their Query-Aware performance. However, using the summarization prompt as a Proxy-Query substantially mitigates this performance loss which demonstrates that the summarization task induces a focus on semantically salient information, making our diagnostic prompt an effective and efficient proxy for a real user query. This finding is particularly valuable for applications where the final query is not available beforehand.
Finally, we note that across all three conditions, SGD-KV consistently outperforms HeadKV, underscoring the robustness and superior design of our summarization-guided, head-level allocation strategy. 
Additional ablation studies are provided on different KV cache budgets in Appendix~\ref{sec:appendix:budget} and on various head configurations in Appendix~\ref{sec:appendix:variants}.
\section{Conclusion}

In this work, we move beyond conventional, retrieval-based heuristics for identifying specialized attention heads by introducing a new functional class: summarization heads. We proposed a novel chunk-summarization diagnostic task to identify and quantify these heads, which are critical for hierarchical information synthesis. Our resulting framework, SGD-KV, integrates this functional understanding into the KV cache management process, setting a new SOTA on complex, long-context benchmarks while reducing memory usage by up to 75\%.
Furthermore, our ablation study indicate that a generic summarization prompt can serve as a highly effective proxy query for token selection, mitigating performance degradation when the final question is unavailable. 
By demonstrating that abstract reasoning roles can be identified and leveraged, our work paves the way for more efficient and interpretable models in the million-token era. 





\bibliographystyle{unsrtnat}
\bibliography{ref.bib}

@article{qwen2.5,
      title={Qwen2.5-1M Technical Report}, 
      author={An Yang and Bowen Yu and Chengyuan Li and Dayiheng Liu and Fei Huang and Haoyan Huang and Jiandong Jiang and Jianhong Tu and Jianwei Zhang and Jingren Zhou and Junyang Lin and Kai Dang and Kexin Yang and Le Yu and Mei Li and Minmin Sun and Qin Zhu and Rui Men and Tao He and Weijia Xu and Wenbiao Yin and Wenyuan Yu and Xiafei Qiu and Xingzhang Ren and Xinlong Yang and Yong Li and Zhiying Xu and Zipeng Zhang},
      journal={arXiv preprint arXiv:2501.15383},
      year={2025}
}

@misc{qwen3technicalreport,
      title={Qwen3 Technical Report}, 
      author={Qwen Team},
      year={2025},
      eprint={2505.09388},
      archivePrefix={arXiv},
      primaryClass={cs.CL},
      url={https://arxiv.org/abs/2505.09388}, 
}

@online{DatabricksBlog2023DollyV2,
    author    = {Mike Conover and Matt Hayes and Ankit Mathur and Jianwei Xie and Jun Wan and Sam Shah and Ali Ghodsi and Patrick Wendell and Matei Zaharia and Reynold Xin},
    title     = {Free Dolly: Introducing the World's First Truly Open Instruction-Tuned LLM},
    year      = {2023},
    url       = {https://www.databricks.com/blog/2023/04/12/dolly-first-open-commercially-viable-instruction-tuned-llm},
    urldate   = {2023-06-30}
}

@inproceedings{dialogsum,
    title = "{D}ialog{S}um: {A} Real-Life Scenario Dialogue Summarization Dataset",
    author = "Chen, Yulong  and
      Liu, Yang  and
      Chen, Liang  and
      Zhang, Yue",
    booktitle = "Findings of the Association for Computational Linguistics: ACL-IJCNLP 2021",
    month = aug,
    year = "2021",
    address = "Online",
    publisher = "Association for Computational Linguistics",
    url = "https://aclanthology.org/2021.findings-acl.449",
    doi = "10.18653/v1/2021.findings-acl.449",
    pages = "5062--5074",
}

@inproceedings{samsum,
    title = "{SAMS}um Corpus: A Human-annotated Dialogue Dataset for Abstractive Summarization",
    author = "Gliwa, Bogdan  and
      Mochol, Iwona  and
      Biesek, Maciej  and
      Wawer, Aleksander",
    booktitle = "Proceedings of the 2nd Workshop on New Frontiers in Summarization",
    month = nov,
    year = "2019",
    address = "Hong Kong, China",
    publisher = "Association for Computational Linguistics",
    url = "https://www.aclweb.org/anthology/D19-5409",
    doi = "10.18653/v1/D19-5409",
    pages = "70--79"
}

@inproceedings{cnndm,
    title = "Get To The Point: Summarization with Pointer-Generator Networks",
    author = "See, Abigail  and
      Liu, Peter J.  and
      Manning, Christopher D.",
    booktitle = "Proceedings of the 55th Annual Meeting of the Association for Computational Linguistics (Volume 1: Long Papers)",
    month = jul,
    year = "2017",
    address = "Vancouver, Canada",
    publisher = "Association for Computational Linguistics",
    url = "https://www.aclweb.org/anthology/P17-1099",
    doi = "10.18653/v1/P17-1099",
    pages = "1073--1083",
}

@article{xsum,
  title={Don't Give Me the Details, Just the Summary! Topic-Aware Convolutional Neural Networks for Extreme Summarization},
  author={Shashi Narayan and Shay B. Cohen and Mirella Lapata},
  journal={ArXiv},
  year={2018},
  volume={abs/1808.08745}
}

@inproceedings{wikilingua, title = "{W}iki{L}ingua: A New Benchmark Dataset for Cross-Lingual Abstractive Summarization", author = "Ladhak, Faisal and Durmus, Esin and Cardie, Claire and McKeown, Kathleen", booktitle = "Findings of the Association for Computational Linguistics: EMNLP 2020", month = nov, year = "2020", address = "Online", publisher = "Association for Computational Linguistics", url = "https://aclanthology.org/2020.findings-emnlp.360", doi = "10.18653/v1/2020.findings-emnlp.360", pages = "4034--4048", }

@article{fu2024not,
  title={Not All Heads Matter: A Head-Level KV Cache Compression Method with Integrated Retrieval and Reasoning},
  author={Fu, Yu and Cai, Zefan and Asi, Abedelkadir and Xiong, Wayne and Dong, Yue and Xiao, Wen},
  journal={arXiv preprint arXiv:2410.19258},
  year={2024}
}

@article{cai2024pyramidkv,
  title={Pyramidkv: Dynamic kv cache compression based on pyramidal information funneling},
  author={Cai, Zefan and Zhang, Yichi and Gao, Bofei and Liu, Yuliang and Liu, Tianyu and Lu, Keming and Xiong, Wayne and Dong, Yue and Chang, Baobao and Hu, Junjie and Xiao Wen},
  journal={arXiv preprint arXiv:2406.02069},
  year={2024}
}

@InProceedings{Gutenberg,
  author    = {Lahiri, Shibamouli},
  title     = {{Complexity of Word Collocation Networks: A Preliminary Structural Analysis}},
  booktitle = {Proceedings of the Student Research Workshop at the 14th Conference of the European Chapter of the Association for Computational Linguistics},
  month     = {April},
  year      = {2014},
  address   = {Gothenburg, Sweden},
  publisher = {Association for Computational Linguistics},
  pages     = {96--105},
  url       = {http://www.aclweb.org/anthology/E14-3011}
}

@misc{bercovich2025llamanemotronefficientreasoningmodels,
      title={Llama-Nemotron: Efficient Reasoning Models}, 
      author={Akhiad Bercovich and Itay Levy and Izik Golan and others},
      year={2025},
      eprint={2505.00949},
      archivePrefix={arXiv},
      primaryClass={cs.CL},
      url={https://arxiv.org/abs/2505.00949}, 
}

@inproceedings{
xiao2025duoattention,
title={DuoAttention: Efficient Long-Context {LLM} Inference with Retrieval and Streaming Heads},
author={Guangxuan Xiao and Jiaming Tang and Jingwei Zuo and junxian guo and Shang Yang and Haotian Tang and Yao Fu and Song Han},
booktitle={The Thirteenth International Conference on Learning Representations},
year={2025},
url={https://openreview.net/forum?id=cFu7ze7xUm}
}

@article{feng2024ada,
  title={Ada-kv: Optimizing kv cache eviction by adaptive budget allocation for efficient llm inference},
  author={Feng, Yuan and Lv, Junlin and Cao, Yukun and Xie, Xike and Zhou, S Kevin},
  journal={arXiv preprint arXiv:2407.11550},
  year={2024}
}

@inproceedings{jiang2024minference,
  author = {Huiqiang Jiang and Yucheng Li and Chengruidong Zhang and Qianhui Wu and Xufang Luo and Surin Ahn and Zhenhua Han and Amir H. Abdi and Dongsheng Li and Chin-Yew Lin and Yuqing Yang and Lili Qiu},
  booktitle = {The Thirty-eighth Annual Conference on Neural Information Processing Systems},
  title = {{MI}nference 1.0: Accelerating Pre-filling for Long-Context {LLM}s via Dynamic Sparse Attention},
  url = {https://openreview.net/forum?id=fPBACAbqSN},
  year = {2024}
}

@misc{gemini25,
      title={Gemini 2.5: Pushing the Frontier with Advanced Reasoning, Multimodality, Long Context, and Next Generation Agentic Capabilities}, 
      author={Gheorghe Comanici and Eric Bieber and Mike Schaekermann and others},
      year={2025},
      eprint={2507.06261},
      archivePrefix={arXiv},
      primaryClass={cs.CL},
      url={https://arxiv.org/abs/2507.06261}, 
}

@inproceedings{
xiao2024sink,
title={Efficient Streaming Language Models with Attention Sinks},
author={Guangxuan Xiao and Yuandong Tian and Beidi Chen and Song Han and Mike Lewis},
booktitle={The Twelfth International Conference on Learning Representations},
year={2024},
url={https://openreview.net/forum?id=NG7sS51zVF}
}

@inproceedings{
zhang2023h2o,
title={H2O: Heavy-Hitter Oracle for Efficient Generative Inference of Large Language Models},
author={Zhenyu Zhang and Ying Sheng and Tianyi Zhou and Tianlong Chen and Lianmin Zheng and Ruisi Cai and Zhao Song and Yuandong Tian and Christopher Re and Clark Barrett and Zhangyang Wang and Beidi Chen},
booktitle={Thirty-seventh Conference on Neural Information Processing Systems},
year={2023},
url={https://openreview.net/forum?id=RkRrPp7GKO}
}

@misc{openai_mrcr,
  author = {OpenAI},
  title = {MRCR},
  year = {2025},
  publisher = {Hugging Face},
  howpublished = {\url{https://huggingface.co/datasets/openai/mrcr}},
  note = {Accessed: [2025-05]}
}

@misc{openai_graphwalks,
  author = {OpenAI},
  title = {GraphWalks},
  year = {2025},
  publisher = {Hugging Face},
  howpublished = {\url{https://huggingface.co/datasets/openai/graphwalks}},
  note = {Accessed: [2025-05]}
}

@inproceedings{babilong,
 author = {Kuratov, Yuri and Bulatov, Aydar and Anokhin, Petr and Rodkin, Ivan and Sorokin, Dmitry and Sorokin, Artyom and Burtsev, Mikhail},
 booktitle = {Advances in Neural Information Processing Systems},
 editor = {A. Globerson and L. Mackey and D. Belgrave and A. Fan and U. Paquet and J. Tomczak and C. Zhang},
 pages = {106519--106554},
 publisher = {Curran Associates, Inc.},
 title = {BABILong: Testing the Limits of LLMs with Long Context Reasoning-in-a-Haystack},
 url = {https://proceedings.neurips.cc/paper_files/paper/2024/file/c0d62e70dbc659cc9bd44cbcf1cb652f-Paper-Datasets_and_Benchmarks_Track.pdf},
 volume = {37},
 year = {2024}
}

@article{ethic,
  title={ETHIC: Evaluating Large Language Models on Long-Context Tasks with High Information Coverage},
  author={Lee, Taewhoo and Yoon, Chanwoong and Jang, Kyochul and Lee, Donghyeon and Song, Minju and Kim, Hyunjae and Kang, Jaewoo},
  journal={arXiv preprint arXiv:2410.16848},
  year={2024}
}

@inproceedings{vllm,
  title={Efficient Memory Management for Large Language Model Serving with PagedAttention},
  author={Woosuk Kwon and Zhuohan Li and Siyuan Zhuang and Ying Sheng and Lianmin Zheng and Cody Hao Yu and Joseph E. Gonzalez and Hao Zhang and Ion Stoica},
  booktitle={Proceedings of the ACM SIGOPS 29th Symposium on Operating Systems Principles},
  year={2023}
}

@inproceedings{zheng2024llamafactory,
  title={LlamaFactory: Unified Efficient Fine-Tuning of 100+ Language Models},
  author={Yaowei Zheng and Richong Zhang and Junhao Zhang and Yanhan Ye and Zheyan Luo and Zhangchi Feng and Yongqiang Ma},
  booktitle={Proceedings of the 62nd Annual Meeting of the Association for Computational Linguistics (Volume 3: System Demonstrations)},
  address={Bangkok, Thailand},
  publisher={Association for Computational Linguistics},
  year={2024},
  url={http://arxiv.org/abs/2403.13372}
}

@misc{dca,
      title={Training-Free Long-Context Scaling of Large Language Models}, 
      author={Chenxin An and Fei Huang and Jun Zhang and Shansan Gong and Xipeng Qiu and Chang Zhou and Lingpeng Kong},
      year={2024},
      eprint={2402.17463},
      archivePrefix={arXiv},
      primaryClass={cs.CL}
}

@article{li2024snapkv,
  title={SnapKV: LLM Knows What You are Looking for Before Generation},
  author={Li, Yuhong and Huang, Yingbing and Yang, Bowen and Venkitesh, Bharat and Locatelli, Acyr and Ye, Hanchen and Cai, Tianle and Lewis, Patrick and Chen, Deming},
  journal={arXiv preprint arXiv:2404.14469},
  year={2024}
}

@inproceedings{dao2023flashattention2,
  title={Flash{A}ttention-2: Faster Attention with Better Parallelism and Work Partitioning},
  author={Dao, Tri},
  booktitle={International Conference on Learning Representations (ICLR)},
  year={2024}
}

@inproceedings{
hsu2025ligerkernel,
title={Liger-Kernel: Efficient Triton Kernels for {LLM} Training},
author={Pin-Lun Hsu and Yun Dai and Vignesh Kothapalli and Qingquan Song and Shao Tang and Siyu Zhu and Steven Shimizu and Shivam Sahni and Haowen Ning and Yanning Chen and Zhipeng Wang},
booktitle={Championing Open-source DEvelopment in ML Workshop @ ICML25},
year={2025},
url={https://openreview.net/forum?id=36SjAIT42G}
}

@inproceedings{deepspeed,
author = {Rasley, Jeff and Rajbhandari, Samyam and Ruwase, Olatunji and He, Yuxiong},
title = {DeepSpeed: System Optimizations Enable Training Deep Learning Models with Over 100 Billion Parameters},
year = {2020},
isbn = {9781450379984},
publisher = {Association for Computing Machinery},
address = {New York, NY, USA},
url = {https://doi.org/10.1145/3394486.3406703},
doi = {10.1145/3394486.3406703},
booktitle = {Proceedings of the 26th ACM SIGKDD International Conference on Knowledge Discovery \& Data Mining},
pages = {3505–3506},
numpages = {2},
location = {Virtual Event, CA, USA},
series = {KDD '20}
}

@inproceedings{
wu2025retrieval,
title={Retrieval Head Mechanistically Explains Long-Context Factuality},
author={Wenhao Wu and Yizhong Wang and Guangxuan Xiao and Hao Peng and Yao Fu},
booktitle={The Thirteenth International Conference on Learning Representations},
year={2025},
url={https://openreview.net/forum?id=EytBpUGB1Z}
}

@inproceedings{
tang2025razorattention,
title={RazorAttention: Efficient {KV} Cache Compression Through Retrieval Heads},
author={Hanlin Tang and Yang Lin and Jing Lin and Qingsen Han and Danning Ke and Shikuan Hong and Yiwu Yao and Gongyi Wang},
booktitle={The Thirteenth International Conference on Learning Representations},
year={2025},
url={https://openreview.net/forum?id=tkiZQlL04w}
}

\newpage
\appendix
\section{Appendix}
\subsection{KV Cache Budget Allocation \& Redistribution}
\label{sec:appendix:redist}

Duo to the task complexity requirements and the goal of minimal performance degradation, we employ a relatively large KV cache budget, which can result in $S_h\cdot L\cdot H\cdot R > 1$ for extremely high-importance heads. 
We address this through a water-filling inspired redistribution algorithm:
\begin{algorithm}[h]
\caption{Summarization-Aware Budget Redistribution}
\textbf{Input:} Normalized head importance scores $S_h$ for all heads $h$,  $M$ is the middle sequence length \\
\textbf{Output:} Redistributed budget allocation $\tilde{b}_h$ for all heads
\begin{algorithmic}[1]
\STATE Initialize $b_h$ for all heads using Equation (6)
\STATE Compute excess budget: $E = \sum{h: b_h > M} (b_h - M)$
\STATE Cap over-allocated heads: $\tilde{b}_h \leftarrow \min(b_h, M)$ for all $h$
\STATE Sort remaining heads by $S_h$ in descending order
\STATE Distribute $E$ to top-scoring heads with available capacity
\end{algorithmic}
\end{algorithm}

This approach prioritizes high-importance summarization heads while maintaining budget constraints, ensuring that critical attention mechanisms receive adequate KV cache allocation for effective long-context summarization.

\subsection{Fine-tuning details}
\label{sec:appendix:sft}
\textbf{Fine-tuning Dataset}. The fine-tuning dataset comprises: 10K synthetic MRCR samples, 20K synthetic OpenAI GraphWalks samples~\citep{openai_graphwalks}, and 25K BABILong fine-tuning samples \citep{babilong}. 
To prevent overfitting, we incorporate 10K samples from the Gutenberg dataset~\citep{Gutenberg} and 10K samples from the Llama-Nemotron-Post-Training-Dataset-v1.1~\citep{bercovich2025llamanemotronefficientreasoningmodels} for regularization. 

\noindent
\textbf{Fine-tuning Recipe}. We use Llama Factory library~\citep{zheng2024llamafactory} as the framework to supervised fine-tune the Qwen2.5-7B-Instruct-1M~\citep{qwen2.5} model. We use full-parameter fine-tuning, cosine learnting schedule with inital learning rate as $1.0\times 10^{-5}$, warmming-up ratios is 0.1, total batch size is 128 and total training epochs is 2. To reduce GPU memory we use flash attention~\citep{dao2023flashattention2}, DeepSpeed with stage 0~\citep{deepspeed} and Liger kernel~\citep{hsu2025ligerkernel}.

\noindent
\textbf{Baselines}. We compare SGD-KV against four strong baselines representing different approaches to KV cache optimization:
(i) DuoAttention~\citep{xiao2025duoattention} finetunes a gate function to binary classify heads into retrieval heads (receiving full KV cache) and streaming heads (retaining only recent tokens and attention sink tokens). Since we do not finetune LLMs on retrieval datasets, we use R2 scores to binary partition heads based on available KV cache budget.
(ii) AdaKV~\citep{feng2024ada} pioneered head-level KV cache budget allocation by analyzing top-K attention values across heads during inference.
(iii) HeadKV~\citep{fu2024not} introduced retrieval-reasoning heads, computing offline importance scores for each head and allocating KV cache budget proportionally based on these scores.
(iv) MInference~\citep{jiang2024minference} determines optimal attention patterns for each head offline and dynamically constructs sparse indices based on assigned patterns during inference.
For fair comparison with DuoAttention, we adapt their binary classification approach by using R2 scores to partition heads into retrieval and streaming categories according to our KV cache budget constraints, rather than performing finetuning. 
For the MInference, since we directly use the VLLM~\citep{vllm} to get the results, it always integrates with the Dual Chunk Attention (DCA)~\citep{dca}

\noindent
\textbf{Configuration. } For all baseline methods, we retain the first 1024 tokens as the attention sink, while maintaining a context window of 1024 recent tokens for the Qwen2.5-7B-1M model. For the larger Qwen3-32B model, both the sink size and window size are reduced to 128 tokens. 
All KV cache allocations occur before repeating the KV values, meaning that the KV cache memory savings are built upon the GQA. 
We also found that if we rerun the identification process for the models after SFT, the HeadKV method with new configuration got worse worse performance. Therefore, we used the configuration of the original checkpoint for HeadKV, DuoAttetnion and our method.

\subsection{Quantitatively Analysis of the Stability and Generalization of the Summarization Heads}
\label{sec:appendix:iou}

At first, we investigate whether summarization heads exhibit consistent patterns across different datasets and tasks. Since we do not impose hard thresholds to classify heads as "summarization heads," we maintain continuous importance scores for all heads, following the approach of HeadKV \citep{fu2024not}.

\begin{figure*}[t]
  \includegraphics[width=0.48\linewidth]{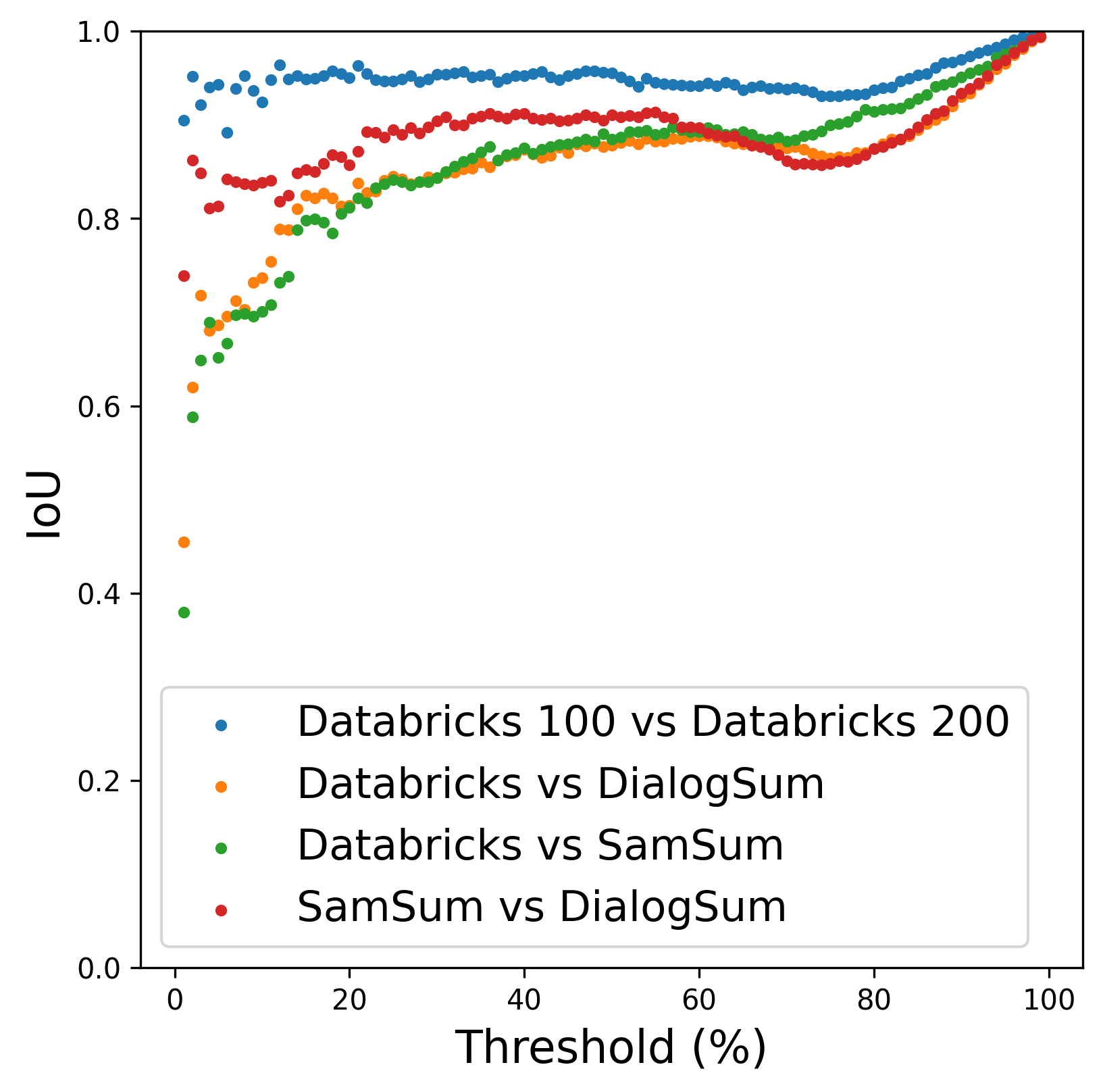} \hfill
  \includegraphics[width=0.48\linewidth]{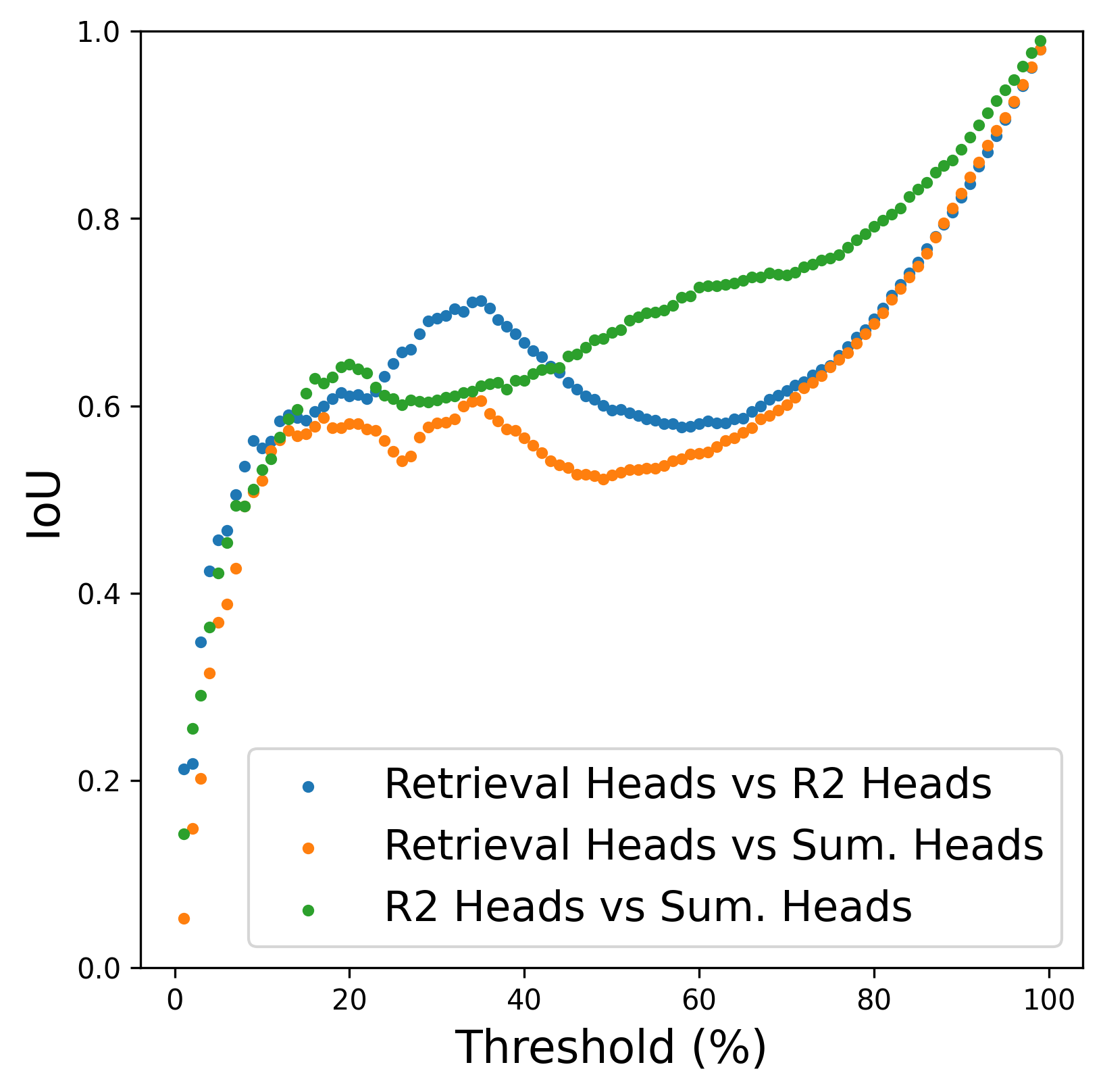}
  \caption {The comparisons of IoU scores between summarization heads with different datasets (left) and different types of heads (right).}
    \label{fig:iou}
\end{figure*}

To quantitatively assess distributional consistency, we rank all attention heads by their importance scores in descending order and compute Intersection over Union (IoU) between ranked lists from different experimental conditions:
\begin{equation}
    \text{IoU}@k = \frac{|\mathcal{H}_1^{(k)} \cap \mathcal{H}_2^{(k)}|}{|\mathcal{H}_1^{(k)} \cup \mathcal{H}_2^{(k)}|}
\end{equation}
where $\mathcal{H}_i^{(k)}$ represents the top-$k$ heads from ranking $i$.

Fig.~\ref{fig:iou} demonstrates high consistency in summarization head identification. The blue curve shows IoU between two disjoint 100-sample subsets from Databricks Dolly dataset \citep{DatabricksBlog2023DollyV2}, achieving near-perfect overlap (IoU > 0.9), indicating robust head identification within datasets.
Cross-dataset analysis reveals substantial generalization: Dolly vs. DialogSum \citep{dialogsum} (orange), Dolly vs. SAMSum (green), and DialogSum vs. SAMSum (red) maintain high IoU scores, particularly for top-ranked heads. Notably, dialogue-based datasets (DialogSum and SAMSum) exhibit stronger similarity, suggesting task-specific head specialization.

For the final configuration of the summarization heads, we use the average scores from the 6 datasets, DialogSum dataset~\citep{dialogsum}, SAMSum dataset~\citep{samsum}, CNN/Dailymail dataset~\citep{cnndm}, Extreme Summarization (XSum) dataset~\citep{xsum}, and the summarization part of the Databricks Dolly dataset~\citep{DatabricksBlog2023DollyV2} and WikiLingua~\citep{wikilingua} dataset.

We also compare summarization heads with retrieval heads and recently proposed R2 heads. The right panel of Fig.~\ref{fig:iou} shows that while the top 20\% of heads overlap significantly across head types, the 20-60\% percentile range reveals distinct preferences, confirming that summarization heads capture unique attention patterns beyond simple retrieval mechanisms.

\subsection{Visualization of Different Types of Heads}
\label{sec:appendix:viz}
\begin{figure*}[h]
  \includegraphics[width=0.98\linewidth]{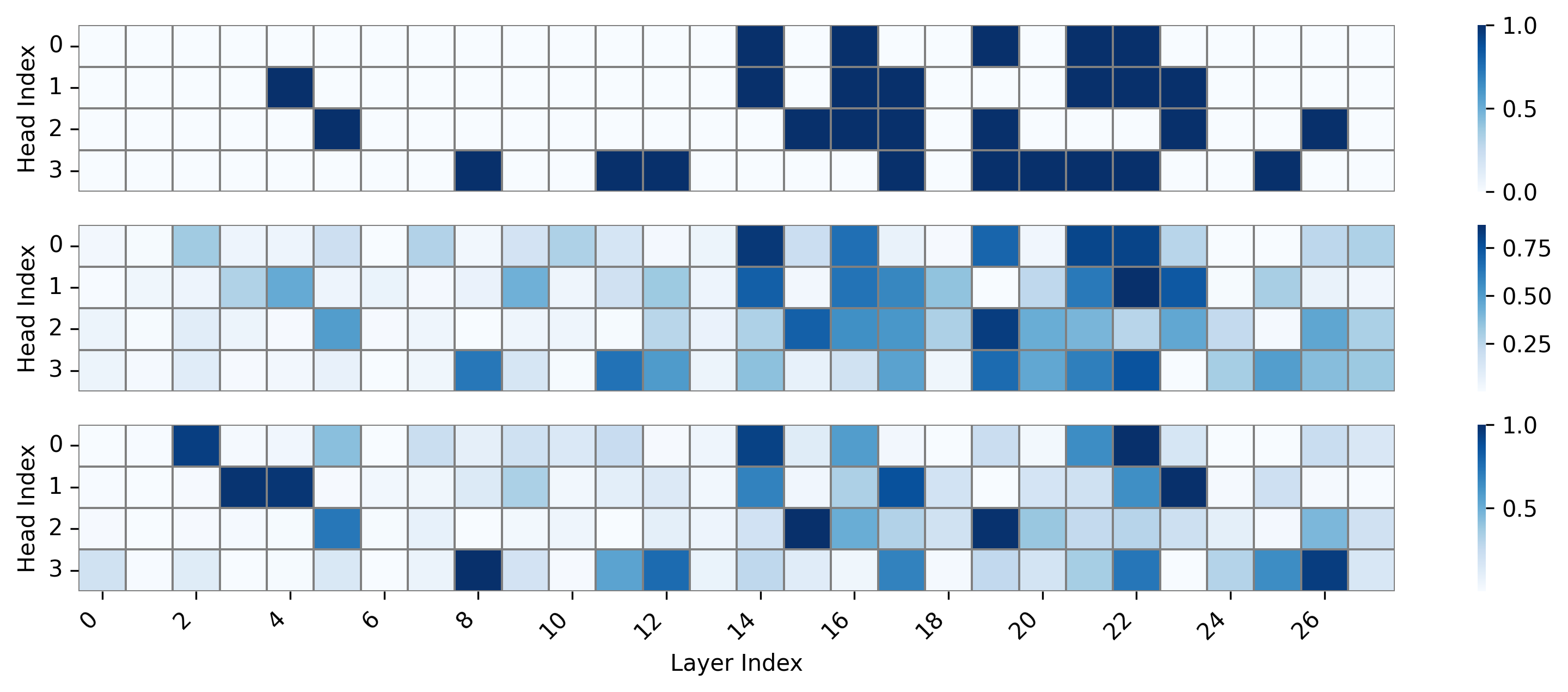}
  \caption {The  visualization of the confusion matrix of the head scores of DuoAttention (top), HeadKV (center) and ours (bottom) in Qwen2.5-7B-Instruct-1M model. Y axis is the index of the heads, 4 in total,  and the X axis is the index of the layers, 28 in total.}
    \label{fig:viz:qwen25}
\end{figure*}

As shown in Fig.~\ref{fig:viz:qwen25}, we visualize the KV cache budget allocation across different attention heads. Since we evict KV cache entries before they repeat, the visualization displays 4 rows representing head indices and 28 columns representing layer indices.
Both R2 heads and summarization heads exhibit similar patterns in their KV cache distributions, indicating that retrieval capability serves as a foundational ability underlying our summarization heads, which encompass both simple retrieval mechanisms and complex information aggregation processes.
There are also few distinct pattern between R2 heads and summarization heads, our method allocates relatively more budget to earlier layers, which are essential for complex reasoning and summarization tasks. This allocation strategy proves crucial because when faced with complex queries, models may be unable to identify all necessary tokens before generation. In such scenarios, methods like HeadKV may inadvertently evict key information that lacks direct surface-level connections to the query but remains semantically relevant for comprehensive understanding.

\subsection{BABILong Results}
\label{sec:appendix:babi}

We also evaluate the methods on the BABILong benchmark~\citep{babilong}{}, which is specifically designed to test the retrieval and reasoning capabilities of LLMs. As shown in Table~\ref{tab:babilong}, despite HeadKV's use of similar question formats to identify R2 heads, it shows negligible differences compared to our summarization heads and achieves similar accuracy to the full KV cache model. This further validates the generalization ability of our summarization heads.
Furthermore, both HeadKV and our method outperform DuoAttention, suggesting that when the KV cache budget is sufficient for head-level KV eviction methods, they typically deliver superior performance compared to DuoAttention methods under the same KV cache constraints.

\begin{table}[h]
  \centering
  \caption{Accuracy comparison between different methods on BABILong dataset (average over QA1 to QA5) from 32K to 1M sequence length using fine-tuned Qwen2.5-7B-Instruct-1M.}
\begin{tabular}{lccccccccc}
\toprule
Method & KV Budget (\%) & 32k & 64k & 128k & 256k & 512k & 1M \\
\midrule
FullKV                  & 100   & 95.4 & 96.6 & 97.2 & 95.6 & 96.2 & 94.6 \\
DAC+Minference          & -     & 95.8 & 96.2 & 97.6 & 96.0 & 87.2 & 76.6 \\
DuoAttention (static)   & 25    & 89.2 & 92.8 &	92.0 & 95.6 & 88.4 & 88.2 \\
HeadKV                  & 25    & 94.8 & 96.0 &	96.6 & 95.8 & 96.0 & 94.2 \\
SGD-KV (Ours)            & 25    & 94.4 & 95.8 &	96.6 & 95.6 & 95.8 & 94.2 \\
\bottomrule
  \end{tabular}
  \label{tab:babilong}
\end{table}

\subsection{Performance under Different KV Cache Budget}
\label{sec:appendix:budget}
\begin{figure}[h]
  \includegraphics[width=0.98\linewidth]{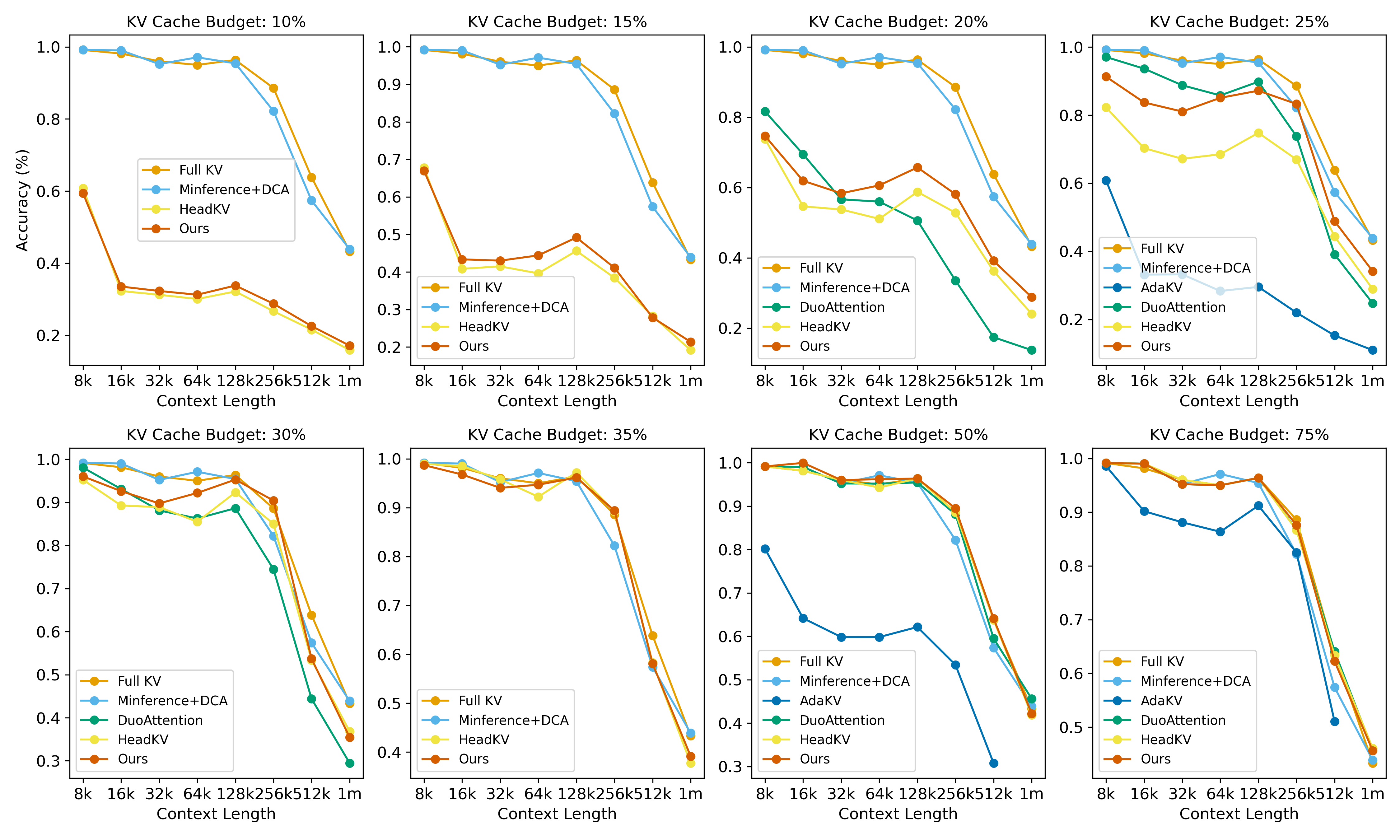} \hfill
  \caption {The comparisons of accuracy on the MRCR dataset under Different KV Cache Budget.}
    \label{fig:mrcr:budget}
\end{figure}
The results in Fig.\ref{fig:mrcr:budget} demonstrate the robust performance of SGD-KV, which consistently surpasses both AdaKV and HeadKV across almost all KV cache budgets.
The performance margin narrows only in extreme cases (KV cache budgets <15\% or >50\%).
Moreover, when the cache budget exceeds 50\%, both SGD-KV and HeadKV begin to outperform MInference, particularly for context lengths longer than 256K. 
This finding aligns with our results on the BABILong benchmark, confirming that given an adequate cache, head-level eviction strategies are more effective than token-level approaches like DuoAttention under identical budget constraints.
Due to time and computational constraints, we omitted some results for DuoAttention (at 10\%, 15\%, 35\% budgets) and AdaKV (at 10\%, 15\%, 20\%, 30\%, 35\% budgets). Nevertheless, the trends established by the available data strongly suggest that these omissions do not alter our overall conclusions.

\subsection{Head Configurations Variants}
\label{sec:appendix:variants}
In this section, we conduct a series of experiments with different configurations of the summarization heads and the R2 heads to evaluate the effectiveness of both types. Due to time and computational resource constraints, we omit some results on the 1M context length. However, since the observed trends are generally consistent across context lengths, the absence of these results does not affect our overall conclusions.
We introduce an additional column in Table~\ref{tab:ablation}, Full KV Heads (\%), which denotes the proportion of attention heads granted access to the full KV cache.

In the first (top) part of Table~\ref{tab:ablation}, we report results for the summarization-head variant of DuoAttention (denoted DuoAttention (Sum.)). Within the DuoAttention framework, using summarization heads consistently underperforms compared to R2 heads. This suggests that R2 heads prioritize ranking heads by retrieval capability, whereas summarization heads primarily focus on fine-grained allocation of KV cache across heads. 
We further evaluate DuoAttention (Sum., Reverse), where the bottom 75\% of heads (with low summarization scores) are assigned full KV cache access, while the top 25\% only retain access to initial and recent tokens. The substantial performance drop demonstrates that constraining the KV cache of high-summarization-score heads is far more detrimental than constraining low-score heads, thereby validating the informativeness of the summarization score.

In the second (middle) part of Table~\ref{tab:ablation}, we present results for SGD-KV (Reverse), where the KV cache allocation is inverted. For instance, a head originally assigned 10\% of the KV cache is instead allocated 90\%. The third row shows that even with $3 \times$ KV cache, SGD-KV (Reverse) performs significantly worse than SGD-KV, further underscoring the importance of score-guided distribution. We also report results for SGD-KV + HeadKV, obtained by averaging the normalized R2 scores and summarization scores. As expected, its performance lies between HeadKV and SGD-KV.

In the last row of the second part, we evaluate SGD-KV (thr), where heads with summarization scores below the mean are set to zero before redistribution. This strategy biases cache allocation toward higher-scoring heads, yielding a larger proportion of full KV heads. Results show improved accuracy for context lengths up to 64k, but degraded performance beyond 128k. This indicates that heads with relatively low summarization scores remain important for ultra-long context modeling.

In the last (bottom) part of Table~\ref{tab:ablation}, we investigate the effect of leveraging summarization scores to guide attention score aggregation in models with GQA.
Take the Qwen2.5-7B-Instruct-1M model as an example: it has 28 attention heads but only 4 key-value (KV) heads, corresponding to a group size of 7. When computing summarization scores, we obtain a score for each individual attention head, but these must be aggregated into a single score per KV head. The most straightforward approaches are averaging or taking the maximum. In our experiments, we adopt the maximum, as it consistently outperforms the mean operation by a small margin.

Based on this, we evaluate three new configurations:
\begin{itemize}
    \item \textbf{Ipt., Max}: multiply the attention scores by the summarization scores for each head, followed by max-pooling within each group.
    \item \textbf{Ipt., Mean}: multiply the attention scores by the summarization scores for each head, followed by mean-pooling within each group.
    \item  \textbf{Ipt., Only}: use only the summarization scores. For instance, if head 3 has the highest summarization score within a group of 7, then all 7 heads inherit the attention scores of head 3.
\end{itemize}

As shown in the results, all three methods improve performance to varying degrees. Among them, Ipt., Max achieves the best overall accuracy from 8k to 512k context lengths, indicating that incorporating summarization scores into attention score aggregation is beneficial for GQA models.

\begin{table*}[htbp]
  \centering
  \caption{Accuracy comparison between different methods on OpenAI MRCR dataset from 8K to 512K sequence length using fine-tuned Qwen2.5-7B-Instruct-1M.}
\begin{tabular}{cccccccccc}
\toprule
Method & \makecell{Full KV\\Heads(\%)} & 8k & 16k & 32k & 64k & 128k & 256k & 512k \\
\midrule
FullKV & 100 & 99.16 & 98.16 & 96.02 & 95.01 & 96.38 & 88.6 & 63.84 \\
\makecell{DuoAttention\\(R2)} & 25 & \textbf{97.09} & \textbf{93.65} & 88.78 & 85.80 & 89.82 & 73.78 & 39.10 \\
\makecell{DuoAttention\\(Sum.)} & 25 & 92.11 & 82.25 & 72.86 & 70.56 & 79.55 & 65.24 & 28.45 \\
\makecell{DuoAttention\\(Sum., Reverse)} & 75 & 45.62 & 14.13 & 12.75 & 8.5 & 5.69 & 5.96 & 6.26 \\
\midrule
HeadKV & 0.0 & 82.27 & 70.31 & 67.18 & 68.52 & 74.81 & 66.98 & 44.36 \\
SGD-KV & 3.57  & 91.30 & 83.73 & 81.06 & 85.09 & 87.19 & 83.29 & 48.86 \\
\makecell{SGD-KV\\(Reverse)} & 0.0 & 47.21 & 17.31 & 15.85 & 11.31 & 8.21 & 8.62 & 8.61 \\
\makecell{SGD-KV\\+ HeadKV} & 1.79 & 88.12 & 77.55 & 75.34 & 75.52 & 82.53 & 77.32 & 45.28 \\
SGD-KV (thr) & 16.96 & 95.51 & 89.62 & 82.49 & 85.43 & 86.24 & 70.03 & 32.22 \\
\midrule
\makecell{SGD-KV\\(Ipt., max)} & 3.57 & 95.5 & 91.49 & \textbf{89.00} & \textbf{87.74} & \textbf{91.16} & \textbf{87.75} & \textbf{57.23} \\
\makecell{SGD-KV\\(Ipt., mean)} & 3.57 & 94.37 & 89.55 & 87.34 & 85.37 & 90.55 & 85.37 & 55.57 \\
\makecell{SGD-KV\\(Ipt., only)} & 3.57 & 95.71 & 86.35 & 86.38 & 87.30 & 90.63 & 82.09 & 49.91 \\
\bottomrule
\end{tabular}
\label{tab:ablation}
\end{table*}

\end{document}